\documentclass[11pt]{article}

\usepackage[final]{acl}

\usepackage{times}
\usepackage{latexsym}

\usepackage[T1]{fontenc}

\usepackage[utf8]{inputenc}

\usepackage{microtype}

\usepackage{inconsolata}

\usepackage{graphicx}

\usepackage{booktabs}
\usepackage{multirow}
\usepackage{amsmath}
 
\usepackage{tcolorbox}
\usepackage{listings}
\usepackage{xcolor}
\usepackage[table]{xcolor} 
\usepackage{graphicx}
\usepackage{amsmath}

\definecolor{promptbg}{RGB}{248,248,248}
\definecolor{promptframe}{RGB}{210,210,210}
\lstdefinestyle{promptstyle}{
  basicstyle=\ttfamily\small,
  columns=fullflexible,
  breaklines=true,
  breakatwhitespace=true,
  keepspaces=true,
  showstringspaces=false,
  frame=none,
  tabsize=2,
  gobble=0,
  breakindent=0pt, 
  breakautoindent=false,
  postbreak=\mbox{} 
}

\newtcolorbox{promptbox}[1][Prompt]{
  colback=promptbg,
  colframe=promptframe,
  title=\textbf{#1},
  boxrule=0.6pt,
  arc=1mm,
  coltitle=black,  
  colbacktitle=gray!25, 
  boxsep=4pt,    
  left=4pt, right=4pt, top=4pt, bottom=4pt 
}

\title{Reflecting Twice before Speaking with Empathy: Self-Reflective Alternating Inference for Empathy-Aware End-to-End Spoken Dialogue}

\author{
 \textbf{Yuhang Jia\textsuperscript{1,2}\thanks{This work was done during an internship at Meituan.}},
 \textbf{Pei Liu\textsuperscript{2}},
 \textbf{Haoqin Sun\textsuperscript{1}},
 \textbf{Jiaming Zhou\textsuperscript{1,2}$^*$},
\textbf{Xuxin Cheng\textsuperscript{2}},
 \\
 \textbf{Cao Liu\textsuperscript{2}},
 \textbf{Ke Zeng\textsuperscript{2}},
 \textbf{Xunliang Cai\textsuperscript{2}},
 \textbf{Yong Qin \textsuperscript{1}\thanks{Corresponding author.}},
\\
 \textsuperscript{1}College of Computer Science, Nankai University\\
 \textsuperscript{2}Meituan LongCat Interaction Team\\
 \small{
   \textbf{Correspondence:} \href{2120240729@mail.nankai.edu.cn}{2120240729@mail.nankai.edu.cn}, 
   \href{qinyong@nankai.edu.cn}{qinyong@nankai.edu.cn}
 }
}

\begin{document}
\maketitle

\begin{abstract}
End-to-end Spoken Language Models (SLMs) hold great potential for paralinguistic perception, and numerous studies have aimed to enhance their capabilities, particularly for empathetic dialogue. However, current approaches largely depend on rigid supervised signals, such as ground-truth response in supervised fine-tuning or preference scores in reinforcement learning. Such reliance is fundamentally limited for modeling complex empathy, as there is no single “correct” response and a simple numerical score cannot fully capture the nuances of emotional expression or the appropriateness of empathetic behavior. To address these limitations, we sequentially introduce \textbf{EmpathyEval}, a descriptive natural-language-based evaluation model for assessing empathetic quality in spoken dialogues. Building upon EmpathyEval, we propose \textbf{ReEmpathy}, an end-to-end SLM that enhances empathetic dialogue through a novel Empathetic Self-Reflective Alternating Inference mechanism, which interleaves spoken response generation with free-form, empathy-related reflective reasoning. Extensive experiments demonstrate that ReEmpathy substantially improves empathy-sensitive spoken dialogue by enabling reflective reasoning, offering a promising approach toward more emotionally intelligent and empathy-aware human-computer interactions.
\end{abstract}
\section{Introduction}

End-to-end Spoken Language Models (SLMs) have experienced significant development in recent years~\cite{defossez2024moshi, zeng2024glm, xu2025qwen2, ding2025kimi, wang2025opens2s, yu2024salmonn,  xu2025qwen3omnitechnicalreport, ai2025ming, team2025longcat, coreteam2025mimoaudioaudiolanguagemodels}, driving substantial progress in conversational AI. Unlike traditional cascaded spoken dialogue systems~\cite{huang2024audiogpt}, which first convert speech into text through automatic speech recognition (ASR), generate responses using large text-based language models (LLMs), and then synthesize the output back into speech (text-to-speech, TTS), end-to-end SLMs inherently support paralinguistic perception. Beyond semantic understanding, they can directly integrate and process a variety of acoustic cues, including emotion, speaker age, gender, and tone (e.g., sarcasm), enabling more contextually appropriate and paralinguistic-aware human-computer interactions.

To fully leverage the paralinguistic perception capabilities of SLMs, several end-to-end optimization approaches have been proposed. These include synthesizing empathetic speech-to-speech data using LLMs combined with controllable TTS for post-training fine-tuning~\cite{geng2025osum, wang2025opens2s}, explicitly aligning acoustic and semantic speech information via rich paralinguistic text annotations~\cite{wang2024blsp, Lu_2025, lu2025desta25audiogeneralpurposelargeaudio}, designing paralinguistic-related training objectives (e.g., emotion classification loss) to facilitate the modeling of acoustic representations~\cite{xue2024chat, wang2025empathy}, and constructing paralinguistic reward models during post-training to better align model behavior with human preferences~\cite{yang2025paras2s}. These methods have been shown to effectively enhance model capabilities in paralinguistic perception and empathetic spoken dialogue. 

However, current approaches, particularly in empathetic dialogue, rely primarily on the rigid end-to-end supervised signals, whether from ground-truth targets or preference scores. Such reliance is problematic for paralinguistic information, \textit{where there is no single “correct” response and a simple assessment score cannot fully capture the nuanced emotional content, speaker intent, or the appropriateness of empathetic expression}, often resulting in insufficient performance improvements.

Considering the inherently complex nature of paralinguistic information, we draw inspiration from reflection mechanisms in LLMs~\cite{madaan2023self, pan2024automatically, bensal2025reflect} and aim to develop a free-form descriptive feedback-based approach for optimizing empathetic spoken dialogue. Accordingly, we first introduce \textbf{EmpathyEval}, a descriptive natural-language-based automatic evaluation model for spoken dialogue empathy. Building upon this evaluation framework, we propose \textbf{ReEmpathy}, an end-to-end spoken language model that enhances empathetic dialogue capabilities through a designed empathetic self-reflective alternating inference mechanism. Leveraging the ability of SLMs to decode additional unspoken tokens alongside speech tokens without incurring extra inference latency~\cite{chiang2025stitch}, ReEmpathy enables seamless free-form natural language self-reflective reasoning during inference, moving beyond the limitations of score-based or single-reference supervised approaches. Our contributions are as follows:

1) We develop \textbf{EmpathyEval}, the first descriptive natural-language-based automatic evaluation model for spoken dialogue empathy, along with a newly constructed and carefully curated dataset of 18,000 Mandarin spoken dialogues, each annotated with a textual description of its empathy quality.

2) We propose \textbf{ReEmpathy}, an end-to-end spoken language model that enhances empathetic dialogue by incorporating a novel empathetic self-reflective alternating inference mechanism, which interleaves response generation with free-form empathetic reflection to promote contextually appropriate and paralinguistic-aware reasoning.

3) Extensive experiments and dedicated ablations demonstrate that ReEmpathy effectively improves contextually appropriate, paralinguistically informed, and empathy-sensitive dialogue, providing a novel perspective beyond conventional score-based or single-reference supervised approaches.

\section{Method}
\subsection{Spoken Empathy Evaluation Dataset}
To support the development of automatic empathy evaluation models, we design a data construction pipeline to generate a speech-to-speech empathetic dialogue dataset annotated with natural-language descriptions of empathetic quality. Inspired by existing approaches for constructing speech-to-speech paralinguistic data~\cite{geng2025osum, wang2025opens2s}, we leverage GPT-4, the zero-shot TTS system CosyVoice2~\cite{du2024cosyvoice}, and the emotion-captioned Chinese seed audio dataset EmotionTalk~\cite{sun2025emotiontalk} to build an automatic data generation pipeline consisting of the following four stages.

\paragraph{Stage 1: Background Construction.}
In preliminary experiments, we found that directly prompting GPT-4 to generate dialogues and emotion labels given only a topic often results in homogeneous, implausible, or insufficiently diverse content (as shown in Table~\ref{tab:perplexity}). To mitigate this issue, we adopt a \textit{One Dialogue, One Story} strategy, in which each dialogue is grounded in a short, coherent narrative describing a specific social situation.

Specifically, we prompt GPT-4 to generate a brief story (approximately five sentences) that depicts a plausible and logically consistent scenario under a given topic, explicitly specifies the relationship between the interlocutors (e.g., family members, colleagues), and describes the main character’s practical needs and emotional state, who subsequently serves as the query speaker in the dialogue. To guide story generation, we consider fifteen common social interaction scenarios, including \textit{Family Chat, Friends Chat, Customer Service, Online Support, Workplace Discussion, Business Negotiation, Media Interaction, Educational Discussion, Healthcare Consultation, Public Service, Dining Service, Travel Assistance, Transportation Inquiry, Financial Consultation, and Emergency Assistance}, ensuring broad coverage of topics.

\paragraph{Stage 2: Query and Response Generation.}  
Based on the generated story, character relationships, and the main character’s needs and emotional state, we prompt a new instance of GPT-4 to generate a user query from the main character’s perspective. Each query is designed to explicitly express the character’s underlying needs and emotional state, which is associated with one of the predefined emotion categories, namely \textit{anger, disgust, fear, happiness, neutral, sadness, and surprise}, following the label taxonomy used in EmotionTalk.

Subsequently, conditioned on the story context and the generated query, the model produces system (assistant) responses with three distinct attitudes: \textbf{Positive}, \textbf{Neutral}, and \textbf{Negative}. 
For Positive responses, the GPT-4 is given full contextual information, including the user’s needs and emotional state, and is instructed to respond empathically. 
For Neutral responses, the GPT-4 has access only to the user’s needs (but no emotional state) and provides objective, literal answers without emotional expression. 
For Negative responses, GPT-4 is has access to all information but generates negative, detached or irrelevant replies while maintaining basic politeness. 
This design produces dialogue pairs that are diverse in attitude, emotionally informative, and contextually grounded.

\paragraph{Stage 3: Empathetic Assessment.} 
In this stage, a separate GPT-4 instance is prompted to evaluate each dialogue’s empathetic quality, conditioned on the story context (character relationships, needs, and emotional state), the user query, and the system response text with emotional annotations. GPT-4 first generates two emotion captions for both the query and response, then assesses the response’s appropriateness from four perspectives: \textit{Need Support (\textbf{NS})}, \textit{Wording Appropriateness (\textbf{WA})}, \textit{Emotion Understanding (\textbf{EU})}, and \textit{Emotional Support (\textbf{ES})}, providing four opinion scores on a 1–5 scale. Following these chain-of-thought reasoning process, GPT-4 ultimately delivers a holistic descriptive empathetic assessment, analyzing the true user intention, evaluating whether the response meets empathetic needs, and offering actionable suggestions to improve empathy where applicable.

\begin{figure}[t]
  \centering
  \includegraphics[width=0.90\columnwidth]{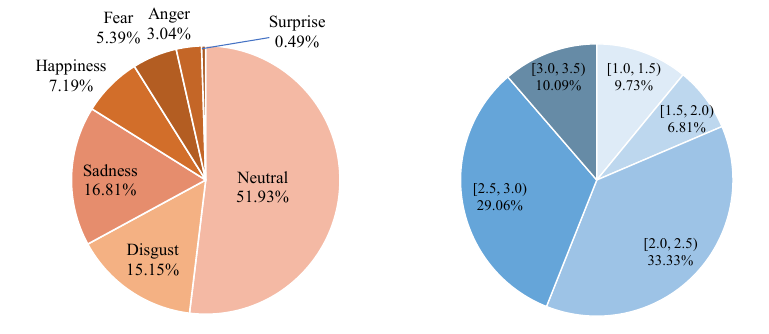}
  \caption{Distribution of dialogue emotion labels and average rating score ranges.}
  \label{fig:pies}
\end{figure}

\begin{figure}[t]
  \centering
  \includegraphics[width=0.95\columnwidth]{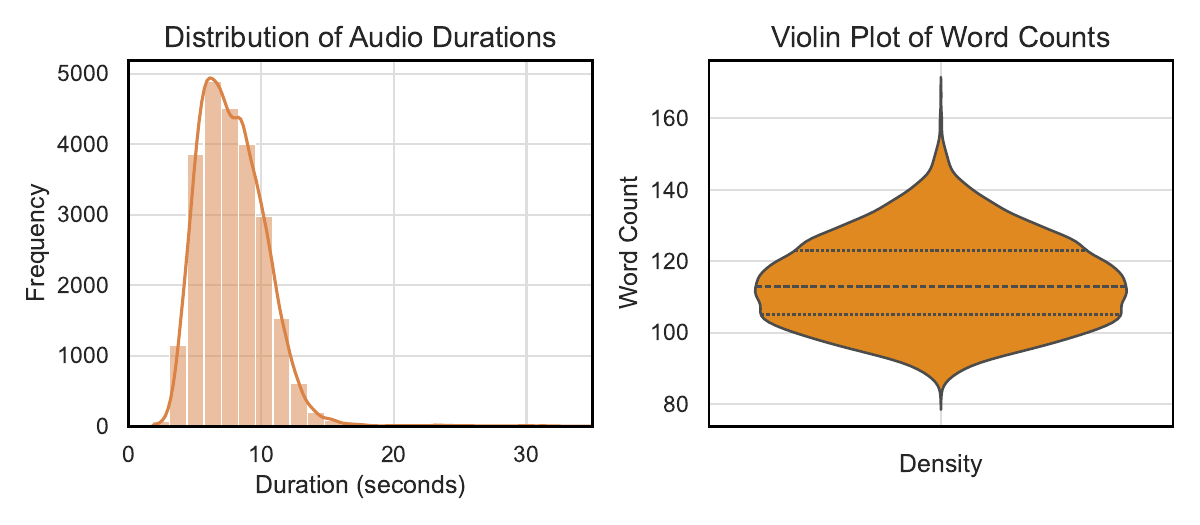}
  \caption{Distribution of dialogue audio durations and assessment word counts.}
  \label{fig:prompt}
\end{figure}

\paragraph{Stage 4: Dialogue Speech Synthesis.} In this stage, we select 140 emotion-captioned seed audios from the EmotionTalk dataset, choosing the highest-confidence samples for each of the seven emotion categories based on five-annotator consensus, and manually filter out multi-speaker or overlapping speech. Using CosyVoice2 with these seed audios, we perform zero-shot text-to-speech synthesis for both user queries and system responses, conditioned on the corresponding emotion and gender labels. The synthesized speech dialogues are then paired with the automatic empathy assessments from Stage~3 to form a fully annotated spoken dialogue empathetic evaluation dataset: 
\begin{equation}
\mathcal{D} = \left\{ 
\left(
x_q^{\text{speech}},\;
x_r^{\text{speech}},\;
E
\right)
\right\}_{i=1}^{N},
\end{equation}

where $x_q^{\text{speech}}$ and $x_r^{\text{speech}}$ denote the spoken user query and system response, respectively, and $E$ represents the descriptive empathy evaluation. Figures~\ref{fig:pies} and~\ref{fig:prompt} respectively illustrate the distributions of labels, scores, and dialogue durations, as well as the lengths of the generated assessment.


\begin{figure*}[t]
  \centering
  \includegraphics[width=1.03\textwidth]{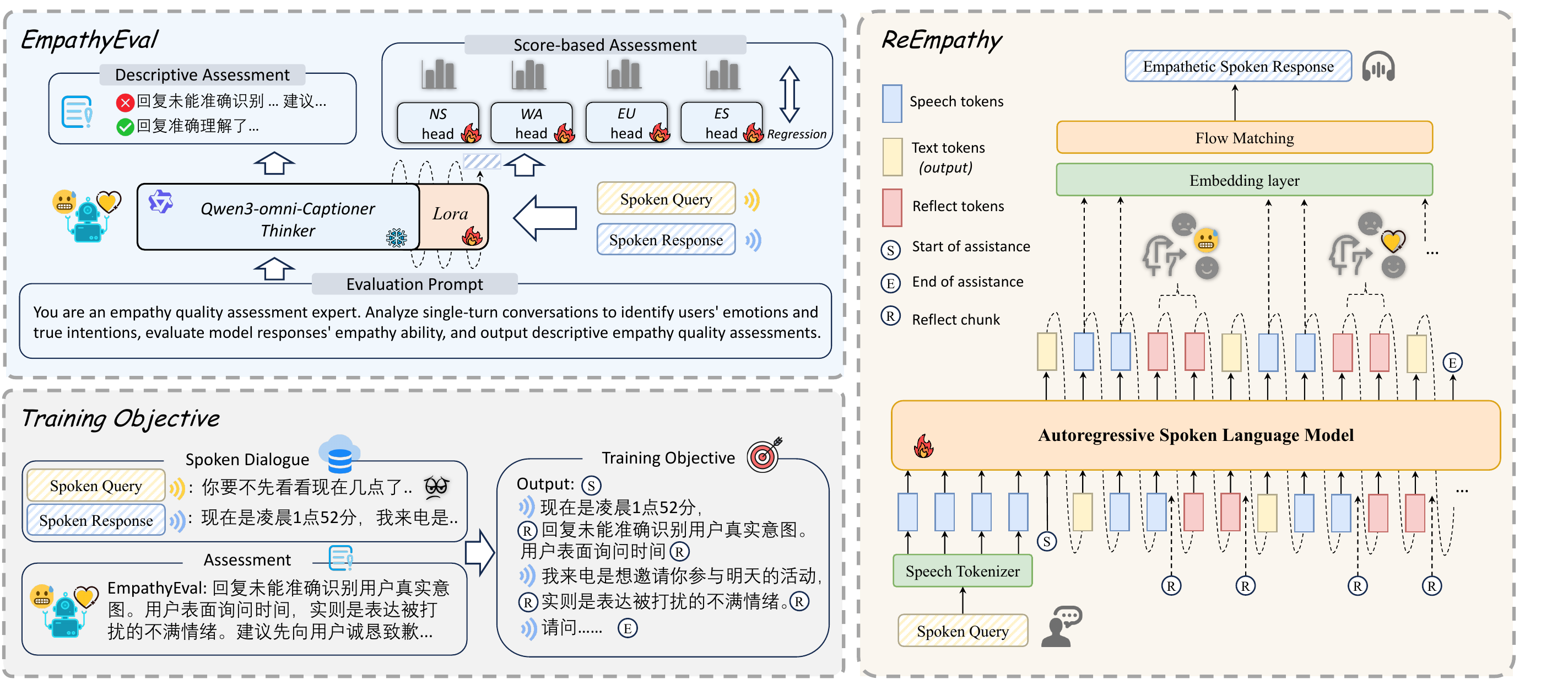}
 \caption{Overview of the proposed EmpathyEval and ReEmpathy framework. 
The upper-left shows the EmpathyEval architecture and its two modes: text-based and score-based assessment. 
The lower-left illustrates the construction of off-policy response--reflection alternating training data. 
The right depicts the alternating inference mechanism of ReEmpathy, where reflective tokens are periodically generated to encourage paralinguistic and
empathetic reasoning.}
\vspace{-2mm}
  \label{fig:reempathy}
\end{figure*}

\subsection{Automatic Evaluation Model}

\begin{table}[h!]
\centering
\small
\begin{tabular}{@{}lccc@{}}
\toprule
Strategies / Perplexity $\downarrow$ & Query  & Response & Dialogue \\
\midrule
wo\_Story                          & 22.16  & 31.23    & 27.99    \\
One Dialogue One Story             & \textbf{19.22}  & \textbf{28.43}    & \textbf{25.59}    \\
\bottomrule
\end{tabular}
\caption{Comparison of GPT-2 perplexity across prompting strategies.}
\label{tab:perplexity}
\end{table}
Based on the constructed spoken dialogue empathy evaluation dataset, we develop our automatic evaluation model on top of Qwen3-Omni-30B-A3B-Captioner, a caption-oriented variant of Qwen3-Omni. To prepare it for empathetic quality assessment, we first adapt Qwen3-Omni to the emotion-related domain by sequentially fine-tuning on the EmotionTalk dataset for speech emotion recognition (SER) and speech emotion captioning tasks, thereby enhancing its capability for paralinguistic emotion perception in Mandarin. We then further fine-tune Qwen3-Omni in a supervised manner using the constructed evaluation dataset to obtain \textbf{EmpathyEval}, which can directly process single-turn spoken dialogues, jointly handling both linguistic content and emotion-related paralinguistic cues. During evaluation, EmpathyEval analyzes the user query intent, assesses the empathetic quality of system responses, and generates a comprehensive, descriptive and natural-language-based evaluation of spoken dialogue empathy in a unified framework.

Building upon EmpathyEval, we further extend Qwen3-Omni by attaching four regression heads on top of its last-layer representation embeddings. These heads are trained to predict the empathy scores along four dimensions, $E_{\textit{NS}}$, $E_{\textit{WA}}$, $E_{\textit{EU}}$, and $E_{\textit{ES}}$ (as defined in Equation~\ref{eq:E_score}), using the 1–5 opinion scores generated during the data construction stage. This yields an effective automatic empathy evaluation tool. Compared with discrete, dialogue-style integer ratings, this continuous, floating-point regression formulation provides finer-grained discrimination and greater sensitivity to subtle differences. The EmpathyEval scoring procedure is defined as follows:
\begin{equation}
E, E_{score} = \text{\textit{EmpathyEval}}\bigl(x_{q}^{\text{speech}}, x_{r}^{\text{speech}}\bigr),
\end{equation}
where
\begin{equation}
E_{score} = \bigl\{E_{\textit{NS}}, E_{\textit{WA}}, E_{\textit{EU}}, E_{\textit{ES}}\bigr\}.
\label{eq:E_score}
\end{equation}

\subsection{Empathetic Self-Reflective Inference}
Building upon EmpathyEval, we propose \textbf{ReEmpathy}, an end-to-end spoken language model that incorporates an empathetic self-reflective alternating inference mechanism. This mechanism forces SLMs to interleave spoken response generation with empathetic self-reflective reasoning, encouraging the model to continuously assess the empathetic quality of its responses, refine them in real time, and generate contextually appropriate, paralinguistic-aware, and empathetic dialogue.

\begin{table*}[t]
  \centering
  \small
\renewcommand{\arraystretch}{0.95}
  \setlength{\tabcolsep}{4pt}
  \begin{tabular}{lcccccccccccc}
    \toprule
    \multirow{3}{*}{Models} & \multicolumn{1}{c}{\textit{SER}} & \multicolumn{10}{c}{\textit{Emotion Captioning $\uparrow$}} \\
    \cmidrule(lr){2-2} \cmidrule(lr){3-12}
                            & Acc\% & Bleu\_1 & Bleu\_2 & Bleu\_3 & Bleu\_4 & Meteor & Rouge\_l & Cider\_d & Spice & Spider & Fense \\
    \midrule
    Qwen2.5-Omni-Ins & 35.58    & 0.2550   & 0.0050   & 0.0000   & 0.0000   & 0.1550  & 0.3447   & 0.0058   & 0.3607 & 0.1833 & 0.8890  \\
    Qwen3-Omni-Cap & 64.26  & 0.1858  & 0.0036  & 0.0000   & 0.0000   & 0.1454 & 0.2885   & 0.0049   & 0.3131 & 0.1590  & 0.8777 \\
    \ \ + Few-shot           & 66.60   & 0.2836  & 0.0057  & 0.0000   & 0.0000   & 0.1268 & 0.3688   & 0.0062   & 0.3779 & 0.1920  & 0.6949 \\
    \ \ + Sft & \textbf{70.75}    & \textbf{0.4268}  & \textbf{0.1820}  & \textbf{0.1305}  & \textbf{0.1070}   & \textbf{0.2242} & \textbf{0.4618}   & \textbf{0.6921}   & \textbf{0.4562} & \textbf{0.5742} & \textbf{0.9356} \\
    \bottomrule
  \end{tabular}
    \caption{Performance improvements of EmpathyEval through SFT on SER and Emotion Captioning tasks.}
  \label{tab:emotion_performance}
\end{table*}

\begin{table*}[t]
  \centering
  \small
\renewcommand{\arraystretch}{0.95}
  \setlength{\tabcolsep}{5pt}
  \begin{tabular}{lcccccccccc}
    \toprule
    \multirow{3}{*}{Models} & \multicolumn{10}{c}{\textit{Empathetic Evaluation $\uparrow$}} \\
    \cmidrule(lr){2-11}
                            & Bleu\_1 & Bleu\_2 & Bleu\_3 & Bleu\_4 & Meteor & Rouge\_l & Cider\_d & Spice & Spider & Fense \\
    \midrule
    Qwen3-Omni-Cap          & 0.3488  & 0.0183  & 0.0052  & 0.0000   & 0.2106 & 0.3997   & 0.0017   & 0.2223 & 0.1120  & \textbf{0.8159} \\
    \ \ + Sft\_Flat          & 0.4807  & 0.1707  & 0.0997  & 0.0537  & 0.3138 & 0.4793   & 0.1922   & \textbf{0.3221} & 0.2571 & 0.7632 \\
    \ \ + Sft\_Grad          & \textbf{0.4860}  & \textbf{0.1745}  & \textbf{0.1025}  & \textbf{0.0057}  & \textbf{0.3176} & \textbf{0.4809}   & \textbf{0.1937}   & 0.3214 & \textbf{0.2575} & 0.7709 \\
    \cmidrule(lr){1-11} 
    ReEmpathy (\textit{F})       & 0.5436  & 0.3245  & 0.2426  & 0.1888   & 0.3796 & 0.5589   & 0.2524   & 0.4504 & 0.8514  & 0.8395 \\
    \bottomrule
  \end{tabular}
    \caption{Performance comparison on the Descriptive Empathetic Evaluation task. \textit{F} is defined in Eq.~\ref{eq:F_score}.}
  \label{tab:empathetic_evaluation_performance}
\end{table*} 

\begin{table*}[t]
  \centering
  \small
\renewcommand{\arraystretch}{0.95}
  \setlength{\tabcolsep}{7pt}
  \begin{tabular}{lcccccccc}
    \toprule
    \multirow{3}{*}{Score Categories} & \multicolumn{4}{c}{GPT4-Pipeline} & \multicolumn{4}{c}{Human MOS} \\
    \cmidrule(lr){2-5} \cmidrule(lr){6-9}
                                      & LCC $\uparrow$ & SRCC $\uparrow$ & KATU $\uparrow$ & MSE $\downarrow$ & LCC $\uparrow$ & SRCC $\uparrow$ & KATU $\uparrow$ & MSE $\downarrow$ \\
    \midrule
    NS (Need Support)            & 0.8520         & 0.8606          & 0.7295          & 0.1830           & 0.7085             & 0.6719              & 0.5085              & 0.7944               \\
    WA (Wording Appropriateness)     & 0.8100         & 0.7812          & 0.6608          & 0.1576           & 0.7036             & 0.6831              & 0.5181              & 0.7854               \\
    EU (Emotion Understanding)       & 0.7832         & 0.7607          & 0.6382          & 0.2237           & 0.6608             & 0.6553              & 0.4896              & 1.1564               \\
    ES (Emotional Support)           & 0.7740         & 0.7412          & 0.6212          & 0.2127           & 0.6721             & 0.6668              & 0.4995              & 1.1288               \\ \cmidrule(lr){1-9}
    \textbf{EmpathyEval (Overall)} & \textbf{0.8585}         & \textbf{0.8629}          & \textbf{0.7235}          & \textbf{0.1209}           & \textbf{0.7125}             & \textbf{0.6940}              & \textbf{0.5256}              & \textbf{0.8090}               \\
    Qwen3-Omni (Original) & 0.6647         & 0.6657          & 0.5377          & 0.6929           & 0.6694             & 0.5936              & 0.4672              & 1.8333               \\
    \bottomrule
  \end{tabular}
    \caption{Correlation results of score-based assessment between EmpathyEval and the GPT pipeline\&Human MOS.}
  \label{tab:gpt4_human_mos_performance}
\end{table*}

\paragraph{Empathetic Self-reflective Inference Mechanism.}
Given a user spoken query $x_q^{\text{speech}}$, ReEmpathy generates the system output as a sequence of fixed-length chunks, alternating between response chunks and reflection chunks. Each response chunk contains both spoken tokens and their corresponding textual transcripts, while reflection chunks consist of internal unspoken reasoning tokens. Specifically, the generation process follows:
\begin{equation}
(r_1, f_1, r_2, f_2, \dots, r_K, f_K) = \text{ReEmpathy}(x_q^{\text{speech}}),
\end{equation}
where $r_k$ denotes the $k$-th response chunk and $f_k$ denotes the corresponding reflective chunk. 

The response and reflection streams are jointly conditioned on the evolving dialogue context, allowing reflective reasoning to influence subsequent response generation at a global level, while the newly generated response continuously informs and shapes the ongoing reflective process. Since we organize the output into fixed-size chunks, individual response and reflection chunks may not be strictly coupled; nevertheless, their collective interactions have been shown to guide the dialogue toward more empathetic behavior. 

The ouput system reply $\hat{x}_r^{\text{speech}}$ and the internal reflection trace $F$ are obtained by concatenation:
\begin{equation}
\hat{x}_r^{\text{speech}} = R = \bigoplus_{k=1}^{K} r_k, \qquad F = \bigoplus_{k=1}^{K} f_k.
\label{eq:F_score}
\end{equation}

\paragraph{Empathetic Self-reflective Supervision Objective.}
To achieve the aforementioned alternating inference behavior, we construct reflective supervision training data based on a large-scale empathetic spoken dialogue dataset. Each dialogue is first evaluated by EmpathyEval to obtain a descriptive assessment. The resulting reflective descriptions are then converted into reflection text tokens and interleaved with the corresponding response audio tokens (and their transcript text tokens) to form the off-policy alternating supervision training data:
\begin{equation}
(x_q^{\text{speech}}, r_1, f_1, r_2, f_2, \dots, r_K, f_K).
\end{equation}

These interleaved sequences are used as training objectives to fine-tune GLM-4-Voice, enabling the model to learn the alternating inference pattern and to integrate reasoning into spoken dialogue generation. Moreover, from a modeling perspective, ReEmpathy not only learns the alternating inference pattern but also produces reflective sequences $F$ that provide accurate empathy assessments, closely matching the results re-obtained from EmpathyEval (see the bottom row of Table~\ref{tab:empathetic_evaluation_performance}).
\begin{equation}
F \sim \text{\textit{EmpathyEval}}\bigl(x_q^{\text{speech}}, \hat{x}_r^{\text{speech}}\bigr).
\end{equation}

Through this design, ReEmpathy employs a novel cross-modal reflective inference mechanism that enables effective empathetic self-reflection. By coupling the response and reflection streams, it has the potential to generate more contextually aware and paralinguistically sensitive dialogue.

\section{Experiments}
\subsection{Datasets}
For tasks involving speech emotion recognition (SER), emotion captioning, and seed audio selection, we use the \textbf{EmotionTalk}~\cite{sun2025emotiontalk} dataset. The dataset is split into training and test sets with a 9:1 ratio and is used to train and validate the model’s ability to perceive emotions in speech.

For the spoken dialogue training benchmark, we adopt the Chinese subset of the \textbf{OpenS2S}~\cite{wang2025opens2s} open-source empathetic dialogue dataset. This subset contains 49,992 speech-to-speech empathetic dialogue pairs and includes annotations for paralinguistic attributes such as emotion, age, and gender. The dataset is split with the first 90\% used for training and the remaining 10\% (indices 44,996--49,994) reserved for evaluation.

For the development and validation of the empathetic assessment model, we use the spoken dialogue empathy evaluation dataset constructed following the procedure described in Section~2.1, split into training and test sets with a ratio of 8:1.

\subsection{Experimental Setup}
\paragraph{Development of EmpathyEval.}
For the first two adaptation stages on the SER and speech emotion captioning tasks, we compare the model’s performance before and after fine-tuning, as well as against Qwen2.5-Omni and the zero-shot setting with five in-domain examples, as reported in Table~\ref{tab:emotion_performance}. For the main assessment task, we compare the model performance before and after fine-tuning and further evaluate the additional gains brought by the staged training strategy by contrasting flat supervised fine-tuning (Sft\_Flat) with gradual multi-stage fine-tuning (Sft\_Grad), as shown in Table~\ref{tab:empathetic_evaluation_performance}.

For the extra score-based assessment, we examine the correlation and accuracy between the predicted scores and the ground-truth scores before and after regression training. The ground-truth scores are constructed using a hybrid strategy: the test split of the training scores is used as large-scale in-domain supervision, and an additional set of 300 human-annotated samples is curated for gold-standard validation. This human-annotated set includes 150 in-domain samples from the constructed dataset and 150 out-of-domain samples selected from the Chinese subset of the OpenS2S empathetic dialogue corpus. Each sample is independently listened and rated by three professional annotators in a blind setting, and the final score is obtained by averaging their ratings.

\paragraph{ReEmpathy.}  To validate the improvement in empathetic capability achieved by ReEmpathy, we adopt the supervised responses from the OpenS2S dialogue dataset as a topline (GroundTruth) and conduct comparative evaluations against several state-of-the-art dialogue models with comparable parameter scales, as well as our baseline model GLM-4-Voice.  Building upon the baseline, we compare four approaches for empathy optimization: 
(1) supervised fine-tuning (SFT) using the speech-to-speech dialogue data; 
(2) direct preference optimization (DPO), where supervised data and model-generated responses are scored by EmpathyEval to construct preference training pairs; 
(3) Chain-of-Thought before Speaking (CoTBS), in which reflective content is generated as a single pre-reasoning step prior to response generation, without iterative alternation; 
and (4) an ablation variant of our method in which reflective reasoning is disabled via instruction.

\begin{figure}[t]
  \centering
  \includegraphics[width=0.60\columnwidth]{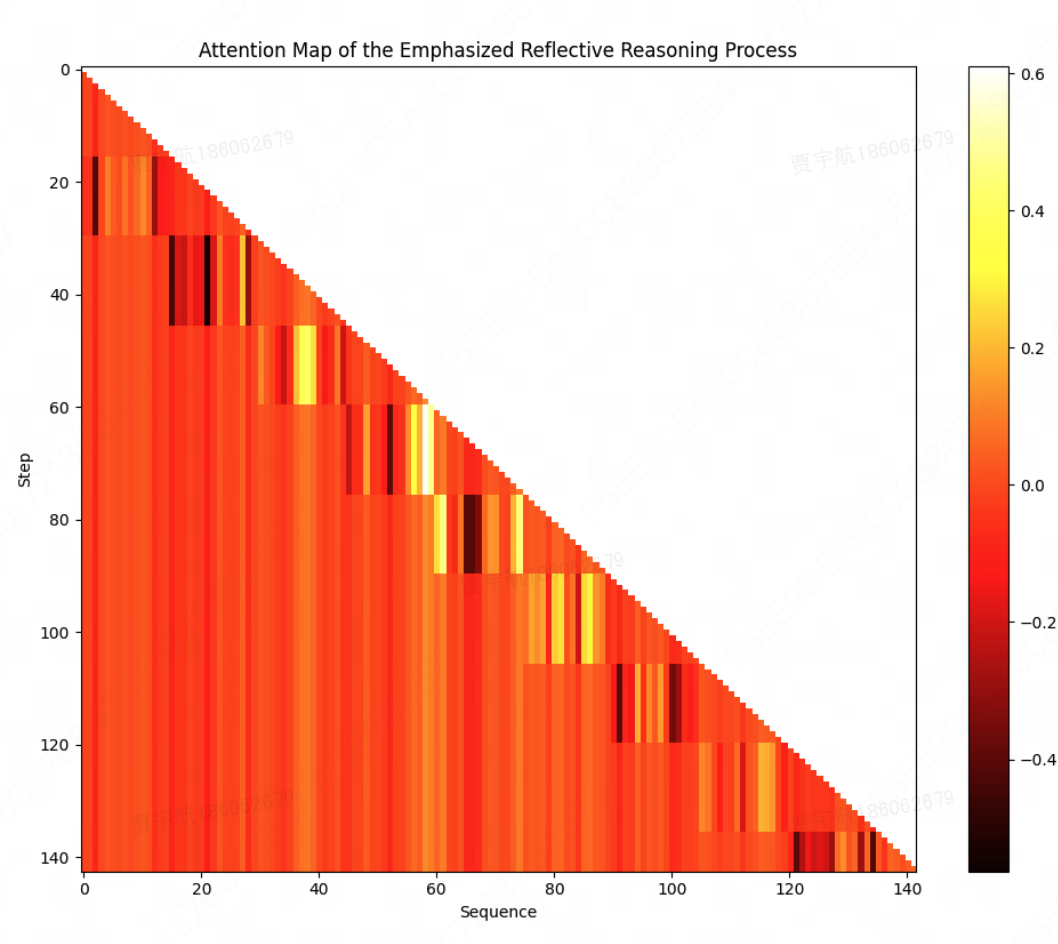}
    \caption{Illustration of adjusting self-attention weights between responses and reflections alternating.}
  \label{fig:attention}
\end{figure}

\begin{table*}[t]
  \centering
  \small
\renewcommand{\arraystretch}{0.95}
  \setlength{\tabcolsep}{6.5pt}
  \begin{tabular}{lcccccccccc}
    \toprule
    \multirow{3}{*}{Models \& Methods} & \multicolumn{5}{c}{EmpathyEval} & \multicolumn{2}{c}{Gpt-4} & \multicolumn{2}{c}{Gpt-4 + Emotion2Vec} \\
    \cmidrule(lr){2-6} \cmidrule(lr){7-8} \cmidrule(lr){9-10}
                      & NS       & WA       & EU       & ES       & Avg $\uparrow$     & AB\_Score $\uparrow$ & MOS $\uparrow$ & AB\_Score $\uparrow$ & MOS $\uparrow$ \\
    \midrule
    GroundTruth       & 3.0495   & 3.5830   & 2.6949   & 2.7291   & 3.0141   & --       & 3.810     & --       & 3.480     \\ \cmidrule(lr){1-10}
    Kimi-Audio        & 2.6604   & 3.2758   & 2.2496   & 2.2026   & 2.5971   & -0.628   & 3.140     & -0.386   & 2.976     \\
    Qwen2.5-Omni       & 2.8379   & 3.3239   & 2.3542   & 2.3691   & 2.7213   & -0.562   & 3.292     & -0.362   & 2.978     \\
    \cmidrule(lr){1-10}
    GLM-4-Voice       & 2.8983   & 3.4300   & 2.4167   & 2.4117   & 2.7892   & -0.330   & 3.616     & -0.458   & 3.308     \\
    \ \ + SFT               & 2.9795   & 3.5758   & 2.6220   & 2.6445   & 2.9555   & -0.096   & 3.734     & -0.156   & 3.402     \\
    \ \ + DPO               & 2.9706   & 3.5657   & 2.6009   & 2.6247   & 2.9405   & -0.132   & 3.726     & -0.120   & 3.356     \\
    \ \ + CoTBS            & 2.9535   & 3.5335   & 2.5594   & 2.5834   & 2.9074   & -0.154   & 3.726     & -0.154   & 3.368     \\
    ReEmpathy (ours)  & \textbf{2.9965}   & \textbf{3.5824}   & \textbf{2.6240}   & \textbf{2.6525}   & \textbf{2.9633}   & \textbf{-0.046}   & \textbf{3.818}     & \textbf{-0.082}   & \textbf{3.430}     \\ 
    \ \ - wo reflect & 2.9744   & 3.5609   & 2.5975   & 2.6244   & 2.9393   & -0.122   & 3.726    & -0.126   & 3.380     \\
    \bottomrule
  \end{tabular}
    \caption{Performance comparison of different models and methods on empathetic dialogue evaluation metrics.}
  \label{tab:empathy_performance}
\end{table*}

Moreover, to further validate the effectiveness of the reflective mechanism, we conduct two sets of quantitative ablation studies. 
In the first set, we gradually and synchronously reduce the chunk size of both reflection and response streams on top of the CoTBS setup, thereby increasing the frequency of reflection. This experiment aims to examine how reflection frequency influences the empathetic quality of the generated outputs.
In the second set, we adjust the self-attention weights between response and reflection chunks under the alternating inference pattern to investigate how mutual information flow between responses and reflections affects empathetic performance. Specifically, as illustrated in Fig.~\ref{fig:attention}, when generating reflection tokens, we allocate more attention to the preceding response content, whereas when generating response tokens, we increase attention to the most recent reflection.

\subsection{Metrics}
For SER, we report classification accuracy (ACC). For emotion captioning and empathy quality evaluation, we adopt standard captioning metrics, including Bleu (Bleu-1 to Bleu-4), Meteor, Rouge-L, Cider-D, Spice, Spider, and Fense, using toolkit\footnote{\url{https://github.com/Labbeti/aac-metrics}}. For score-based assessment, we evaluate the accuracy of the predicted scores using Mean Squared Error (MSE) and report correlation metrics, including the Linear Correlation Coefficient (LCC), Spearman’s Rank Correlation Coefficient (SRCC), and Kendall’s Tau (KTAU)~\cite{cooper2022generalization}.

\begin{figure}[t]
  \includegraphics[width=1.0\columnwidth]{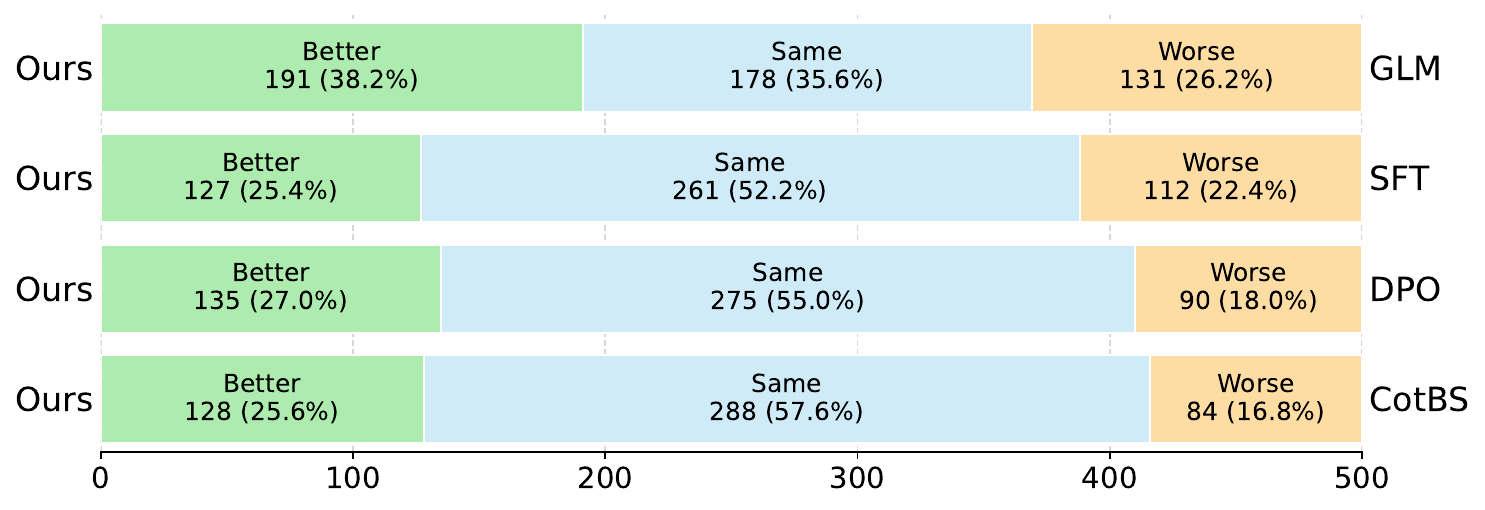} 
    \caption{A/B test results comparing ReEmpathy with other optimization approaches with GPT-4 + E2V.}
  \label{fig:ABtest}
\end{figure}

For dialogue empathetic quality evaluation, we first use EmpathyEval to obtain scores along four dimensions (NS, WA, EU, and ES) and report their averaged values. In addition, we transcribe the dialogue speech using ASR and prompt GPT-4 to evaluate empathy quality. Following~\cite{geng2025osum}, we further incorporate emotion labels predicted by Emotion2Vec-Large\footnote{\url{https://huggingface.co/emotion2vec}} as auxiliary references for GPT-4. With GPT-4 serving as the judge, we conduct A/B testing between the model outputs and the ground-truth responses, assigning +1 if the model output is preferred, -1 if it is worse, and 0 if indistinguishable. We also prompt GPT-4 to provide a 1--5 MOS-style empathy rating. Together, these metrics provide a comprehensive evaluation of empathetic dialogue quality.

\section{Results and Analysis}
\begin{figure*}[t]
  \centering
  \includegraphics[width=0.41\linewidth]{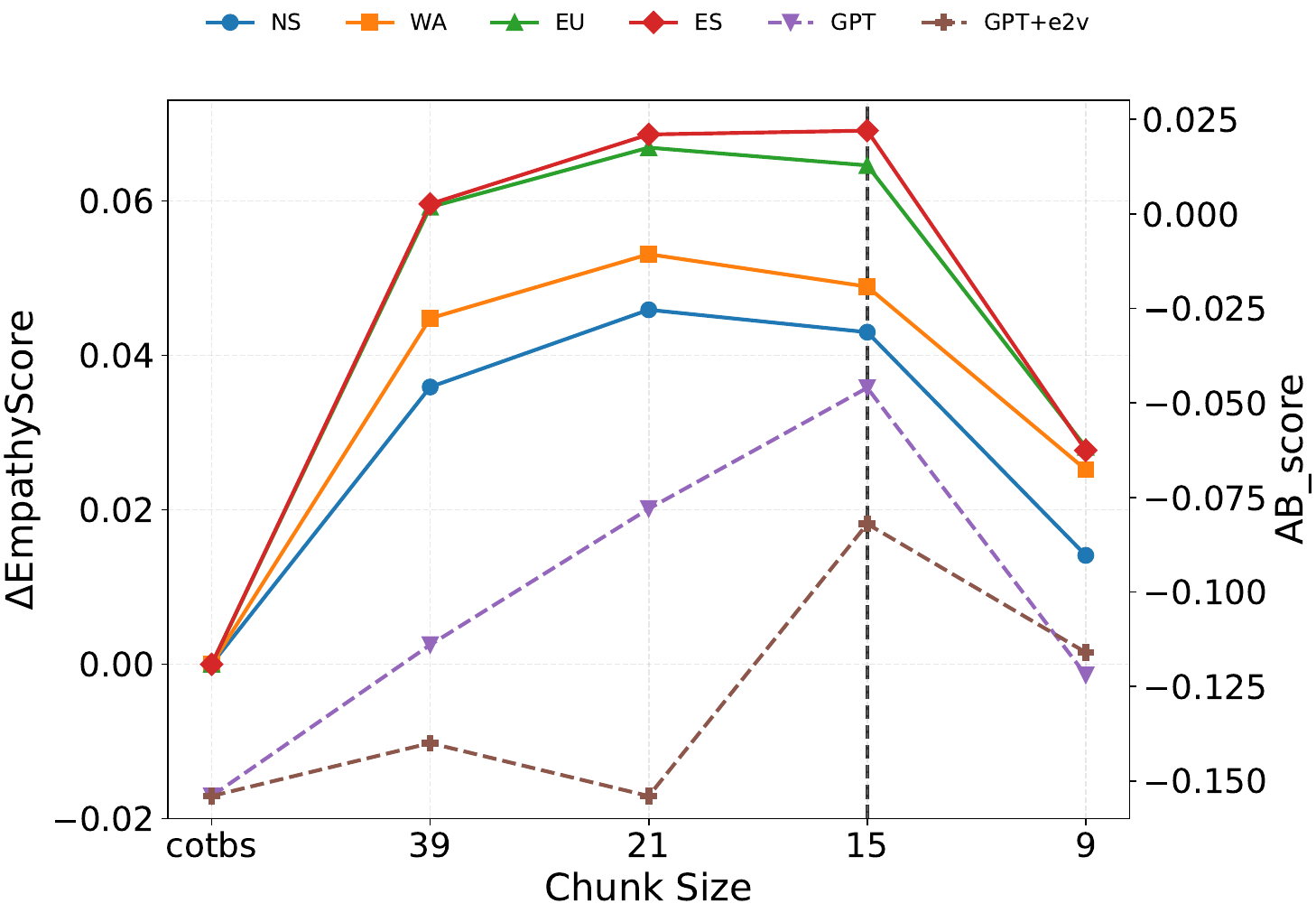}
  \includegraphics[width=0.41\linewidth]{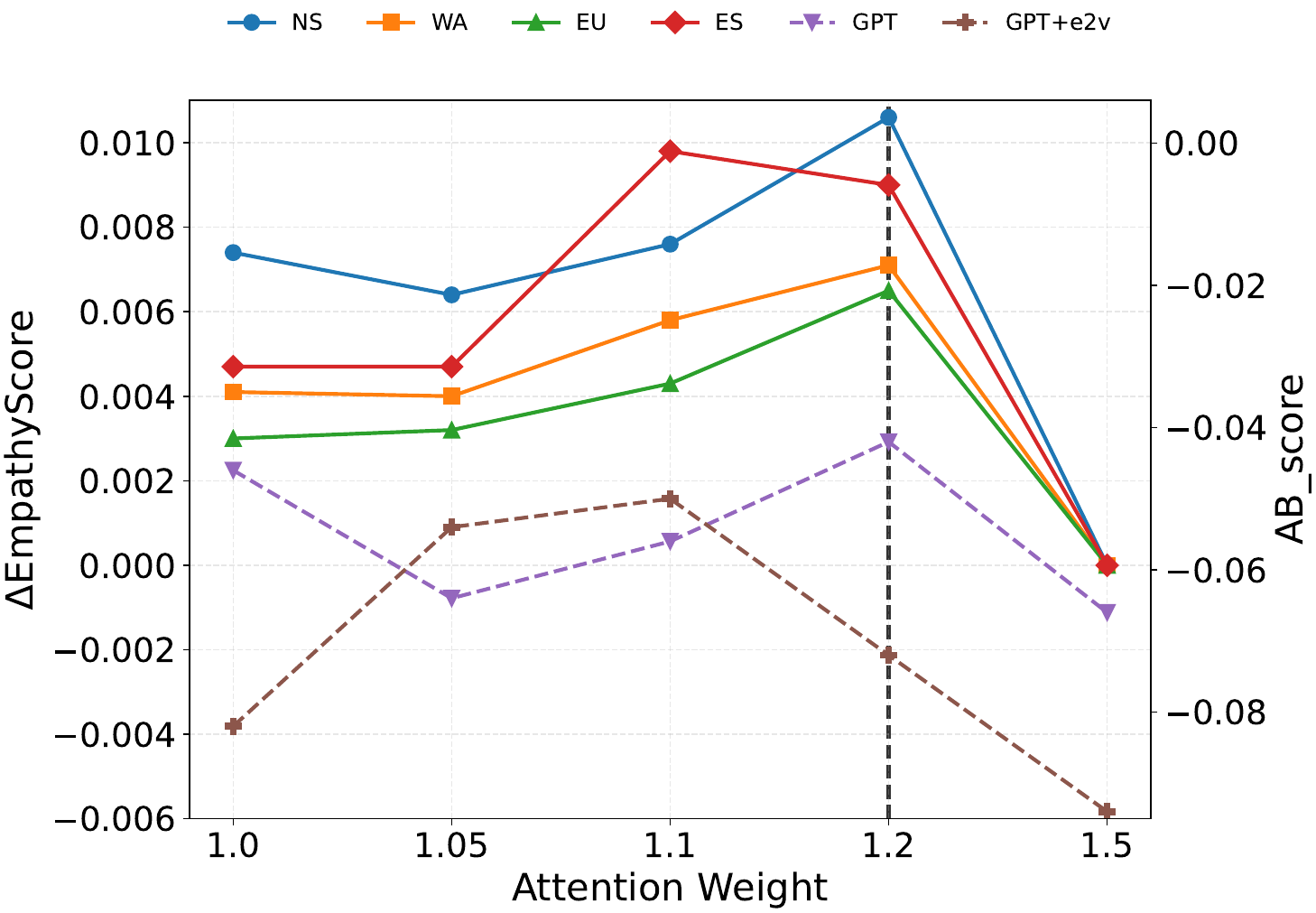}
  \caption {Quantitative ablation results on the impact of alternation frequency and attention weight scaling on empathetic performance.}
  \label{fig:Ablation}
\end{figure*}

\subsection{Performance of EmpathyEval}

As shown in Tables~\ref{tab:emotion_performance} and~\ref{tab:empathetic_evaluation_performance}, after multi-stage fine-tuning, EmpathyEval achieves substantial improvements over the backbone Qwen3-Omni in emotion recognition, emotion captioning, and the quality of descriptive empathy assessment. On the descriptive assessment test set, EmpathyEval attains notable performance, with an average BLEU-1--4 score of 0.20 and a METEOR score of 0.31.
For score-based quantitative evaluation, Table~\ref{tab:gpt4_human_mos_performance} shows that EmpathyEval achieves very high consistency on the test set, with an average LCC of 0.85 and an MSE of 0.12 on the 1–5 scale. On the human-annotated test set, EmpathyEval also maintains strong agreement with human judgments. Although the average MSE increases to 0.81, which is reasonable given the inherent variability in human ratings, the average LCC remains as high as 0.71, indicating substantial consistency with human preferences, particularly in out-of-domain setting.

\subsection{Performance of ReEmpathy}
As shown in Table~\ref{tab:empathy_performance}, in the comparative evaluation of different SLMs, GLM-4-Voice demonstrates an inherent advantage among models of similar size. We attribute this advantage to its speech-style-controlled spoken training design, which more effectively captures paralinguistic characteristics in Chinese spoken dialogues. Among the various approaches for optimizing empathetic dialogue performance, direct end-to-end supervised fine-tuning (SFT) on annotated data proves to be the simplest yet most effective method, outperforming both direct preference optimization (DPO) and Chain-of-Thought (CoT) based approaches. Notably, although CoT introduces empathy-related information as explicit reasoning, it does not yield performance gains and even significantly decreases performance compared with SFT.

Among these approaches, ReEmpathy achieves the best performance across all metrics. Both Table~\ref{tab:empathy_performance} and the A/B test results in Fig.~\ref{fig:ABtest} demonstrate the performance gains introduced by our method. Notably, in the ablation study where reflective inference is disabled, these gains disappear, further confirming that the observed improvements in empathetic capability are attributable to the reflective inference mechanism rather than to the mere inclusion of additional training knowledge.

\subsection{Ablation Study}
As shown in Figure~\ref{fig:Ablation}, we observe that within a reasonable range (approximately $[15, \infty)$), gradually reducing the chunk size in each alternation round, which increases the alternation frequency between response and reflection, leads to improvements in empathetic dialogue quality. However, when the chunk size becomes too small, for example below 9, performance begins to degrade, likely due to excessively fragmented attention and disrupted contextual coherence. For the attention reweighting ablation, we observe that the empathetic quality improves as the amplification factor of the cross-attention between responses and reflections is progressively increased from the original setting ($\text{reWeight}=1.0$), reaching a peak around $\text{reWeight} \in [1.1, 1.2]$. Similarly, further increasing the weight, for example beyond 1.5, results in a decline in performance.

These two sets of ablation results are consistent with our hypotheses regarding alternating generation and reflective reasoning, providing quantitative evidence that supports both the effectiveness and the theoretical validity of the proposed empathetic self-reflective alternating inference mechanism.

\section{Related Work}
\subsection{Paralinguistic-aware Speech Dialogue}
Early efforts to enhance paralinguistic understanding in SLMs primarily focused on speech-to-text dialogue. In these approaches, paralinguistic attributes such as age, gender, emotion, and acoustic environment are explicitly annotated and fed into text LLMs to generate paralinguistically informed textual responses. These responses are then distilled into speech models, enabling SLMs to better align with paralinguistic acoustic information~\cite{wang2024blsp, Lu_2025, lu2025desta25audiogeneralpurposelargeaudio, wang2025crossmodalknowledgedistillationspeech}. Beyond explicit annotation and distillation, another line of work enhances paralinguistic representation by introducing auxiliary training objectives, such as incorporating emoting classification losses as additional supervision signals to explicitly strengthen the modeling of paralinguistic information~\cite{xue2024chat, wang2025empathy}. In end-to-end SLMs, a more prevalent strategy is to construct supervised speech-to-speech empathetic dialogue datasets and perform end-to-end training during the later stages of pre-training or post-training, thereby implicitly enhancing both paralinguistic perception and empathetic dialogue capabilities~\cite{geng2025osum, wang2025opens2s}.

The work most closely related to ours is ParaS2S~\cite{yang2025paras2s}, which constructs a score-based paralinguistic evaluation pipeline using text LLMs, ASR models, and audio reasoning models, and distills Qwen2.5-Omni as a reward model to optimize Kimi-Audio via GRPO~\cite{shao2024deepseekmath}. While both ParaS2S and our approach adopt an evaluation-and-feedback optimization paradigm, our work differs in that we focus on descriptive, text-based empathy assessment and leverage free-form reflective reasoning for empathetic optimization, going beyond conventional reinforcement learning methods.

\section{Conclusion}
In this work, we address the core challenges of enhancing empathetic capabilities in SLMs. We first introduce \textbf{EmpathyEval}, the first descriptive automatic evaluation model for providing fine-grained and comprehensive assessments of spoken dialogue empathy. Building upon this evaluation framework, we propose \textbf{ReEmpathy}, an end-to-end spoken language model that incorporates a novel empathetic self-reflect mechanism, interleaving response generation with free-form empathetic reflection to effectively improve empathy-sensitive dialogue performance. ReEmpathy offers a new perspective beyond conventional supervised fine-tuning and reinforcement learning approaches for empathetic dialogue in end-to-end SLMs, paving a promising pathway toward more emotionally intelligent human-computer interactions.

\section*{Limitations}
While ReEmpathy demonstrates promising improvements in empathetic dialogue through self-reflective alternating inference, several limitations warrant further investigation. First, the coupling between response and reflection streams is currently supervised only at the global dialogue level. At the chunk level, the interaction between individual response and reflection segments cannot yet be trained on-policy, which limits the interpretability and fine-grained control of the alternating inference process. Exploring more granular, on-policy supervision for chunk-level interactions represents a valuable direction for future research. 

Second, the current supervision of the reflective reasoning stream is limited to supervised fine-tuning. Further investigation into reinforcement learning or other advanced optimization strategies for reflective reasoning could potentially enhance the model’s empathetic capabilities.

\section*{Ethics Considerations}
All human annotations were performed by professional annotators who were compensated above the local average rate, under informed consent regarding the purpose and use of the data. The constructed dataset does not contain any personally identifiable information and was processed following privacy-preserving practices.

All resources developed in this work, including the dataset, model checkpoints, training protocols, and code, are under the MIT License if released. AI assistance was used in a responsible and controlled manner. To the best of our knowledge, this work does not pose any foreseeable ethical risks.

\bibliography{custom}

\appendix
\section{Details of the GPT-as-Judge Prompts}
\label{sec:appendixA}

This section presents the prompt templates used for the GPT-as-Judge paradigm, which are used to quantitatively evaluate the empathetic capability of dialogue systems on a 1–5 scale. The templates are illustrated in Fig.~\ref{fig:example_prompt1}, Fig.~\ref{fig:example_prompt2} and Fig.~\ref{fig:example_prompt3}. The A/B test follows the same evaluation criteria, outputs scores of 1, 0, or -1, and alternates the presentation order of A and B to mitigate position bias.

\begin{figure}[ht]
    \centering
    \begin{promptbox}[Example (Translate from Chinese)]
    \begin{lstlisting}[style=promptstyle]
You are a professional single-turn dialogue empathy assessment expert. Based on four inputs---user query, user emotion label, system response, and system response emotion label---you are required to conduct an objective and quantitative evaluation of the empathy quality of the system response.

Please strictly follow the definitions of the four evaluation dimensions below, then synthesize your judgment to output an integer score ranging from 1 to 5 (output only an Arabic numeral, with no additional text whatsoever):
    
Evaluation Dimension Definitions
1.  Need-Support
- High-score criteria (4--5 points): Precisely identifies the user's core needs; the response is highly relevant to the needs and provides substantive answers or solutions.
- Low-score criteria (1--2 points): Completely deviates from the user's needs; gives irrelevant answers, mechanical responses, or perfunctory replies.

2.  Wording-Appropriateness
- High-score criteria (4--5 points): The language is natural and fluent, with clear logic and appropriate etiquette; free of ambiguity or offensive expressions.
- Low-score criteria (1--2 points): The sentences are stiff and obscure, with confused logic; contains semantic misunderstandings or inappropriate wording.
3.  Emotion-Understanding
    \end{lstlisting}
    \end{promptbox}
    \caption{Examples of GPT-MOS Prompt (I)}
    \label{fig:example_prompt1}
\end{figure}

\begin{figure}[t]
  \includegraphics[width=1.01\columnwidth]{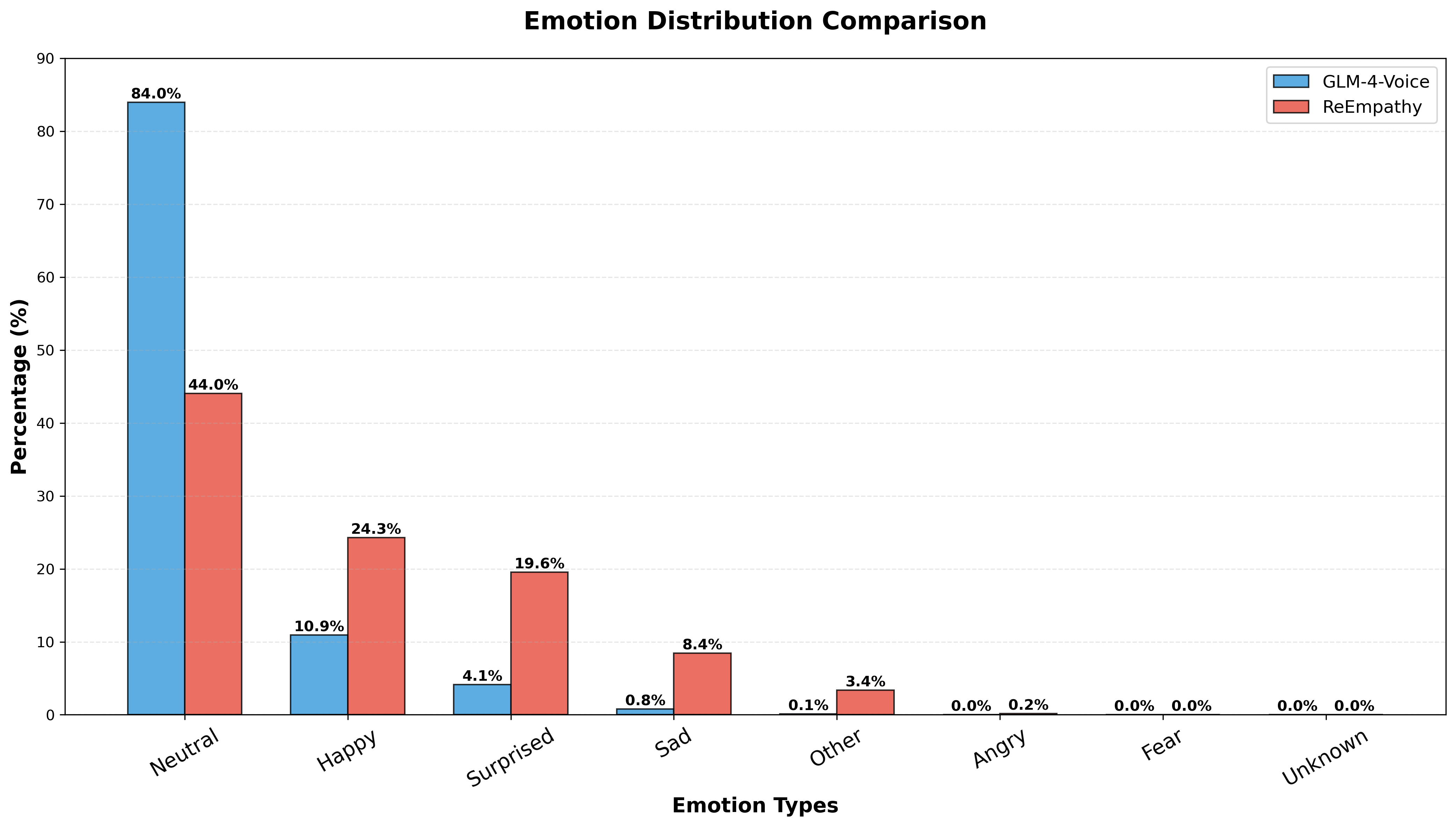} 
    \caption{Comparison of emotion type distribution.}
  \label{fig:EmotionChart}
\end{figure}

\begin{figure}[ht]
    \centering
    \begin{promptbox}[Example (Translate from Chinese)]
    \begin{lstlisting}[style=promptstyle]
- High-score criteria (4--5 points): Accurately recognizes
 the user's explicit emotions and implicit emotional demands; the response demonstrates in-depth insight into the user's emotional state.
- Low-score criteria (1--2 points): Misjudges the user's emotional tendency; completely ignores emotional information or has obvious misunderstandings of emotional demands.

4.  Emotional-Support
- High-score criteria (4--5 points): The response conveys empathic care (e.g., comfort, encouragement, resonance), effectively alleviating the user's negative emotions or reinforcing positive emotions.
- Low-score criteria (1--2 points): Adopts an indifferent and alienated attitude; provides no emotional support at 
    \end{lstlisting}
    \end{promptbox}
    \caption{Examples of GPT-MOS Prompt (II)}
    \label{fig:example_prompt2}
\end{figure}

\begin{figure}[ht]
    \centering
    \begin{promptbox}[Example (Translate from Chinese)]
    \begin{lstlisting}[style=promptstyle]
all, or even exacerbates the user's negative feelings.
Comprehensive Scoring Scale (1--5 Points)
- 5 points: High-emotional-intelligence empathy. Precisely meets user needs + professional wording + in-depth emotional understanding + strong emotional support; perfectly fits the dialogue scenario.
- 4 points: High-quality empathy. Adequately addresses user needs + appropriate wording + accurate emotional understanding + effective emotional support; no obvious shortcomings.
- 3 points: Acceptable empathy. The response is basically relevant to user needs + no obvious wording issues + superficial emotional understanding + insufficient emotional support; no negative impact.
- 2 points: Weak empathy. Deviated response to needs / stiff wording + shallow emotional understanding; no obvious negative effects.
- 1 point: Negative empathy. Completely deviates from user needs + incorrect emotional understanding + inappropriate wording; causes negative impact on the user.

---
Input Information
User Query: \{query\}
User Emotion Label: \{query\_emo\}

System Response: \{model\_resp\}
System Response Emotion Label: \{resp\_emo\}
---

Your Score:
    \end{lstlisting}
    \end{promptbox}
    \caption{Examples of GPT-MOS Prompt (III)}
    \label{fig:example_prompt3}
\end{figure}

\section{Details of Ablation Study Results}
\label{sec:appendixB}

This section reports the detailed ablation results on reflection frequency (as shown in Table\ref{tab:ablation_interleaving}) and attention weight adjustment(as shown in Table\ref{tab:ablation_attention_strength}). All evaluation metrics are computed following the same protocol as in Table~\ref{tab:empathy_performance}.

\begin{table*}[t]
  \centering
  \small
  \renewcommand{\arraystretch}{0.95}
  \setlength{\tabcolsep}{4.5pt}
  \begin{tabular}{lcccccccccc}
    \toprule
    \multirow{3}{*}{Models \& Methods} & \multicolumn{5}{c}{EmpathyEval} & \multicolumn{2}{c}{Gpt-4} & \multicolumn{2}{c}{Gpt-4 + Emotion2Vec} \\
    \cmidrule(lr){2-6} \cmidrule(lr){7-8} \cmidrule(lr){9-10}
                      & NS       & WA       & EU       & ES       & Avg $\uparrow$ & AB\_Score $\uparrow$ & Mos\_Score $\uparrow$ & AB\_Score $\uparrow$ & Mos\_Score $\uparrow$ \\
    \midrule
    CoTBS            & 2.9535   & 3.5335   & 2.5594   & 2.5834   & 2.9074   & -0.154   & 3.726     & -0.154   & 3.368     \\
    Chunk\_09    & 2.9676   & 3.5587   & 2.5875   & 2.6111   & 2.9312   & -0.122   & 3.746     & -0.116   & 3.386     \\
    Chunk\_15    & 2.9965   & 3.5824   & 2.6240   & 2.6525   & 2.9633   & -0.046   & 3.818     & -0.082   & 3.430     \\
    Chunk\_21    & 2.9994   & 3.5866   & 2.6263   & 2.6520   & 2.9661   & -0.078   & 3.748     & -0.154   & 3.396     \\
    Chunk\_39    & 2.9894   & 3.5783   & 2.6186   & 2.6430   & 2.9573   & -0.114   & 3.740     & -0.140   & 3.370     \\
    \bottomrule
  \end{tabular}
  \caption{Ablation study results on reflection frequency.}
  \label{tab:ablation_interleaving}
\end{table*}

\begin{table*}[t]
  \centering
  \small
  \renewcommand{\arraystretch}{0.95}
  \setlength{\tabcolsep}{4.5pt}
  \begin{tabular}{lcccccccccc}
    \toprule
    \multirow{3}{*}{Models \& Methods} & \multicolumn{5}{c}{EmpathyEval} & \multicolumn{2}{c}{Gpt-4} & \multicolumn{2}{c}{Gpt-4 + Emotion2Vec} \\
    \cmidrule(lr){2-6} \cmidrule(lr){7-8} \cmidrule(lr){9-10}
                      & NS       & WA       & EU       & ES       & Avg $\uparrow$ & AB\_Score $\uparrow$ & Mos\_Score $\uparrow$ & AB\_Score $\uparrow$ & Mos\_Score $\uparrow$ \\
    \midrule
    Original         & 2.9965   & 3.5824   & 2.6240   & 2.6525   & 2.9633   & -0.046   & 3.818     & -0.082   & 3.430     \\
    Atten*1.05       & 2.9955   & 3.5823   & 2.6242   & 2.6525   & 2.9636   & -0.064   & 3.796     & -0.054   & 3.414     \\
    Atten*1.1        & 2.9967   & 3.5841   & 2.6253   & 2.6576   & 2.9659   & -0.056   & 3.780     & -0.050   & 3.400     \\
    Atten*1.2        & 2.9997   & 3.5854   & 2.6275   & 2.6568   & 2.9674   & -0.042   & 3.788     & -0.072   & 3.400     \\
    Atten*1.5        & 2.9891   & 3.5783   & 2.6210   & 2.6478   & 2.9591   & -0.066   & 3.790     & -0.094   & 3.420     \\
    \bottomrule
  \end{tabular}
  \caption{Ablation study results on attention reweight.}
  \label{tab:ablation_attention_strength}
\end{table*}

\section{Computational and Training Setup}
\label{sec:appendixC}
For ReEmpathy, fine-tuning was conducted using the LLaMA-Factory framework with the same hyperparameter settings as in~\cite{chiang2025stitch}. The fine-tuning of EmpathyEval was performed using the ms-swift framework with LoRA. The main hyperparameters included a learning rate of $1\times 10^{-4}$, LoRA rank of 8, LoRA alpha of 32, and the adapters were applied to all linear modules. Both models were trained on 8 A100 GPUs, while evaluation was performed on a single A100 GPU.

\begin{table*}[t!]
  \centering
  \small
\renewcommand{\arraystretch}{0.95}
  \setlength{\tabcolsep}{7pt}
  \begin{tabular}{lcccccccc}
    \toprule
    \multirow{3}{*}{Score Categories} & \multicolumn{4}{c}{In Domain} & \multicolumn{4}{c}{Out of Domain} \\
    \cmidrule(lr){2-5} \cmidrule(lr){6-9}
                                      & LCC $\uparrow$ & SRCC $\uparrow$ & KATU $\uparrow$ & MSE $\downarrow$ & LCC $\uparrow$ & SRCC $\uparrow$ & KATU $\uparrow$ & MSE $\downarrow$ \\
    \midrule
    NS (Need Support)            & 0.7807         & 0.7831          & 0.6183          & 0.5777           & 0.3925             & 0.3333              & 0.2451              & 1.0112               \\
    WA (Wording Appropriateness)     & 0.7548         & 0.7822          & 0.6074          & 0.8673           & 0.4276             & 0.3795              & 0.2814              & 0.7035               \\
    EU (Emotion Understanding)       & 0.7099         & 0.6916          & 0.5276          & 0.8052           & 0.4212             & 0.3832              & 0.2833              & 1.5076               \\
    ES (Emotional Support)           & 0.7297         & 0.7316          & 0.5595          & 0.7703           & 0.4025             & 0.3686              & 0.2704              & 1.4873               \\ \cmidrule(lr){1-9}
    \textbf{EmpathyEval (Overall)} & 0.7734         & 0.7821          & 0.6123          & 0.5874           & 0.4320             & 0.3855              & 0.2822              & 1.0306               \\
    \bottomrule
  \end{tabular}
    \caption{Correlation results between EmpathyEval and Human MOS on indomain \& outofdomain.}

  \label{tab:human_mos_details}
\end{table*}

\begin{table*}[t]
  \centering
  \small
  \renewcommand{\arraystretch}{0.95}
  \setlength{\tabcolsep}{4.5pt}
  \begin{tabular}{lcccccccccc}
    \toprule
    \multirow{3}{*}{Models \& Methods} & \multicolumn{5}{c}{EmpathyEval} & \multicolumn{2}{c}{Gpt-4} & \multicolumn{2}{c}{Gpt-4 + Emotion2Vec} \\
    \cmidrule(lr){2-6} \cmidrule(lr){7-8} \cmidrule(lr){9-10}
                      & NS       & WA       & EU       & ES       & Avg $\uparrow$ & AB\_Score $\uparrow$ & Mos\_Score $\uparrow$ & AB\_Score $\uparrow$ & Mos\_Score $\uparrow$ \\
    \midrule
    Atten*1.2 (0-20)       & 2.9992   & 3.5842   & 2.6272   & 2.6559   & 2.9666   & -0.0460   & 3.7900     & \textbf{-0.0600}   & \textbf{3.4180}      \\
    Atten*1.2 (10-30)        & \textbf{2.9997}   & \textbf{3.5854}   & \textbf{2.6275}   & \textbf{2.6568}   & \textbf{2.9674}   & \textbf{-0.0420}   & \textbf{3.7880}     & -0.0720   & 3.4000     \\
    Atten*1.2 (20-40)        & 2.9903   & 3.5781   & 2.6208   & 2.6485   & 2.9594   & -0.0840   & 3.7800     & -0.1060   & 3.4040     \\
    \bottomrule
  \end{tabular}
  \caption{Attention reweight on different transformer layers.}
  \label{tab:ablation_attention_layer}
\end{table*}

\section{Details of the HumanMOS Correlation}
\label{sec:appendixD}
The statistical data for in-domain and out-of-domain datasets, along with the results of linear correlation and mean error for human annotation scores, are shown in Table~\ref{tab:human_mos_details}. These results demonstrate the performance of EmpathyEval in assessing empathetic quality across different domains.

\section{Attention Reweighting Across Different Transformer Layers}
\label{sec:appendixE}
We investigate the performance improvement differences brought by adjusting attention weights at various layers of the Transformer model. As shown in Table~\ref{tab:ablation_attention_layer}, the most effective and robust attention weight adjustments occur at the middle layers (layers 10-20, the setting used in the ablations).

\section{Output Emotion Type Distribution}
\label{sec:appendixF}
We investigate the distribution of output speech emotion types using Emotion2Vector-plus-large. As illustrated in Figure~\ref{fig:EmotionChart}, the optimized model demonstrates a significantly enhanced diversity in emotion distribution.

\end{document}